\documentclass[runningheads]{llncs}
\usepackage[T1]{fontenc}
\usepackage{multirow}
\usepackage{booktabs}
\usepackage{amsfonts}

\usepackage{amsmath}

\usepackage{graphicx}
\usepackage{hyperref}
\begin{document}
\title{FOCUS: Decoupling Expert Personas in LLMs to Enhance Domain Expert Capabilities}
%
%
\makeatletter
\newcommand{\printfnsymbol}[1]{%
  \textsuperscript{\@fnsymbol{#1}}%
}
\makeatother

\author{
Guanyu Wang\inst{1}$^\ast$ \and
Zidi Zhang\inst{2}$^\ast$ \and
Xu Chu\inst{1}$^\dagger$
}
%
\authorrunning{Wang et al.}
%

\institute{
Peking University, Beijing, China\\
\email{\{wgy2023, chuxu\}@stu.pku.edu.cn}
\and
The University of Sydney, Sydney, Australia\\
\email{zzha0350@uni.sydney.edu.au}\\
\textsuperscript{*}Equal contribution \hspace{1em} \textsuperscript{$\dagger$}Corresponding author
}

\maketitle              
\begin{abstract}
Large Language Models (LLMs) can exhibit diverse personas, and activating expert personas has been shown to improve domain expertise and task accuracy. However, existing persona control methods often suffer from cross-domain coupling, which may lead to overly aggressive behavior in high-caution domains such as healthcare, or excessive conservatism in risk-sensitive domains such as financial trading. To address this issue, we propose FOCUS (\textbf{\underline{F}}ine-tuning with \textbf{\underline{O}}rthogonal \textbf{\underline{C}}ontrol for \textbf{\underline{U}}ncoupled persona\textbf{\underline{S}}). FOCUS first automatically extracts expert persona vectors from LLMs, then applies orthogonal decomposition to decouple domain-specific expert personas, and finally introduces an expert gating module to adaptively control persona activation according to task contexts. With a two-stage training strategy and a gated selection regularizer, the model learns to activate appropriate personas for both single-domain and cross-domain tasks. Experiments on financial, legal, medical, and cross-domain benchmarks show that FOCUS improves task accuracy and outperforms existing persona control methods. Our code is available at \href{https://anonymous.4open.science/r/openpersona-48F4}{this url}.

\keywords{LLMs  \and Persona vectors \and Domain experts.}
\end{abstract}
\section{Introduction}

\begin{figure}[htbp]
\centering
\includegraphics[width=0.99\textwidth]{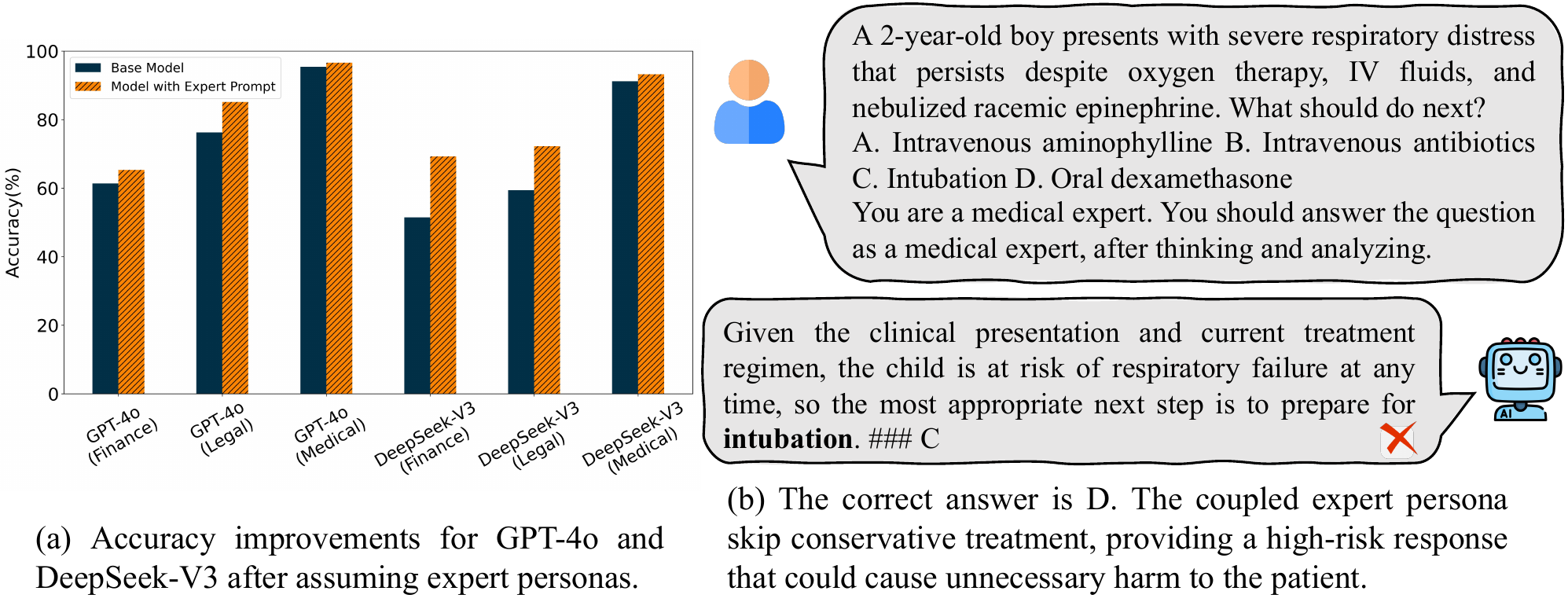}
\caption{Performance of prompting LLMs to assume expert personas on financial, legal, and medical tasks.}
\label{fig:intro}
\end{figure}

Large Language Models (LLMs) can exhibit diverse personas in their generated responses~\cite{chen2025persona}. These personas may manifest as undesirable behaviors, such as threats or unsafe actions~\cite{perrigo2023bing,zhou-etal-2024-making}, or as beneficial traits, such as objectivity and agreeableness~\cite{bai2022constitutional,liu2025prosocial}. Prior studies have shown that assigning expert personas to LLMs can improve downstream task performance, including mathematical reasoning~\cite{yang2023large} and healthcare applications~\cite{tseng2024two}, as illustrated in Fig.~\ref{fig:intro}(a). Existing work further suggests that such personas can be activated through both prompt-based methods~\cite{jiang-etal-2024-personallm} and training-based approaches~\cite{thakurpersonas,kannan2025selfsupervised,chu2025qwen}.

However, the persona boundaries induced by existing expert persona activation methods are often ambiguous, which may lead to behavioral conflicts across domains. This issue arises because different domains impose potentially conflicting requirements on decision-making behavior. For example, the medical domain prioritizes extreme caution~\cite{belisle2024we}, whereas financial trading often requires balancing risk and opportunity~\cite{xiao2024tradingagents}. As a result, activating a coupled expert persona may introduce undesirable cross-domain interference. For instance, as shown in Fig.~\ref{fig:intro}(b), an overly aggressive expert persona, if inadvertently activated in medical QA, may lead to overconfident responses and potential safety risks. Therefore, although domain-specific expert personas can be explicitly activated~\cite{yang2023large,tseng2024two}, their behavior may still be affected by cross-domain coupling in pre-training and post-training data, leading to impure domain alignment and unstable performance.

In this paper, we propose FOCUS (\textbf{\underline{F}}ine-tuning with \textbf{\underline{O}}rthogonal \textbf{\underline{C}}ontrol for \textbf{\underline{U}}ncoupled persona\textbf{\underline{S}}), a persona-vector editing framework designed to decouple and control domain-specific expert personas. FOCUS first uses an automated pipeline to extract expert persona vectors from LLMs, and then applies orthogonal decomposition to decouple domain-specific persona vectors for finance, law, and medicine. Next, we introduce a lightweight expert gating module into the feed-forward network of the LLM and train it to activate the corresponding persona vectors according to the task context. To encourage sparse vector activation in single-domain tasks and induce more focused expert behavior, we add a gated selection regularizer to the training loss. We further design a two-stage training strategy to support both single-domain specialization and cross-domain generalization. Experiments on financial, legal, medical, and cross-domain tasks show that FOCUS improves the accuracy of domain-task answers, offering a new approach to efficient domain adaptation for LLMs. Our contributions are: (a) we identify that coupled expert personas can be unstable on domain tasks due to potential conflicts; (b) we propose FOCUS, which decouples and trains an expert gating module to adaptively activate expert personas; (c) FOCUS improves LLM answer accuracy across multiple domains and cross‑domain tasks.

\section{Related Work}
\subsection{Representation Steering and Editing in LLMs}
Controlling model behavior through manipulation of internal representations has recently emerged as a promising alternative to parameter fine-tuning \cite{rw01}. Prior work on activation steering and representation engineering shows that modifying hidden states can effectively influence model outputs without updating model weights \cite{rw02,rw03}. These approaches include inference-time interventions, linear direction-based editing, and activation patching \cite{rw04}.

Closely related to this line of work are subspace-based methods, such as concept erasure and null space projection techniques, which aim to remove specific attributes by projecting representations onto orthogonal subspaces \cite{rw05,rw06}. These methods highlight that different semantic factors can be disentangled within the representation space \cite{rw07,wang2025learn}.

Our method builds upon these insights but differs in both objective and design. Instead of removing unwanted attributes, we aim to isolate domain-specific behavioral factors and make them controllable.

\subsection{Multi-domain Adaptation in LLMs}
Adapting large language models to multiple domains is a long-standing challenge \cite{rw11,tan2026rado,wang2026muse,chu2025graphsos}. Existing approaches typically rely on multi-task learning or instruction tuning, where models are trained on data from diverse domains to improve generalization \cite{rw12,rw13}. While these methods can enhance overall performance, they rely on implicit knowledge sharing and do not provide explicit control over domain-specific behaviors.

As a result, when different domains impose conflicting requirements, such as conservative decision-making in medical applications versus risk-sensitive reasoning in financial scenarios, models may exhibit unstable or inconsistent behavior \cite{rw14,chu2026towards}. This limitation becomes more pronounced in cross-domain settings, where multiple domain-specific factors must be considered simultaneously.

Our work provides a complementary perspective by introducing structured control over domain-specific representations.

\section{Methodology}

\begin{figure}[htbp]
\centering
\includegraphics[width=0.99\textwidth]{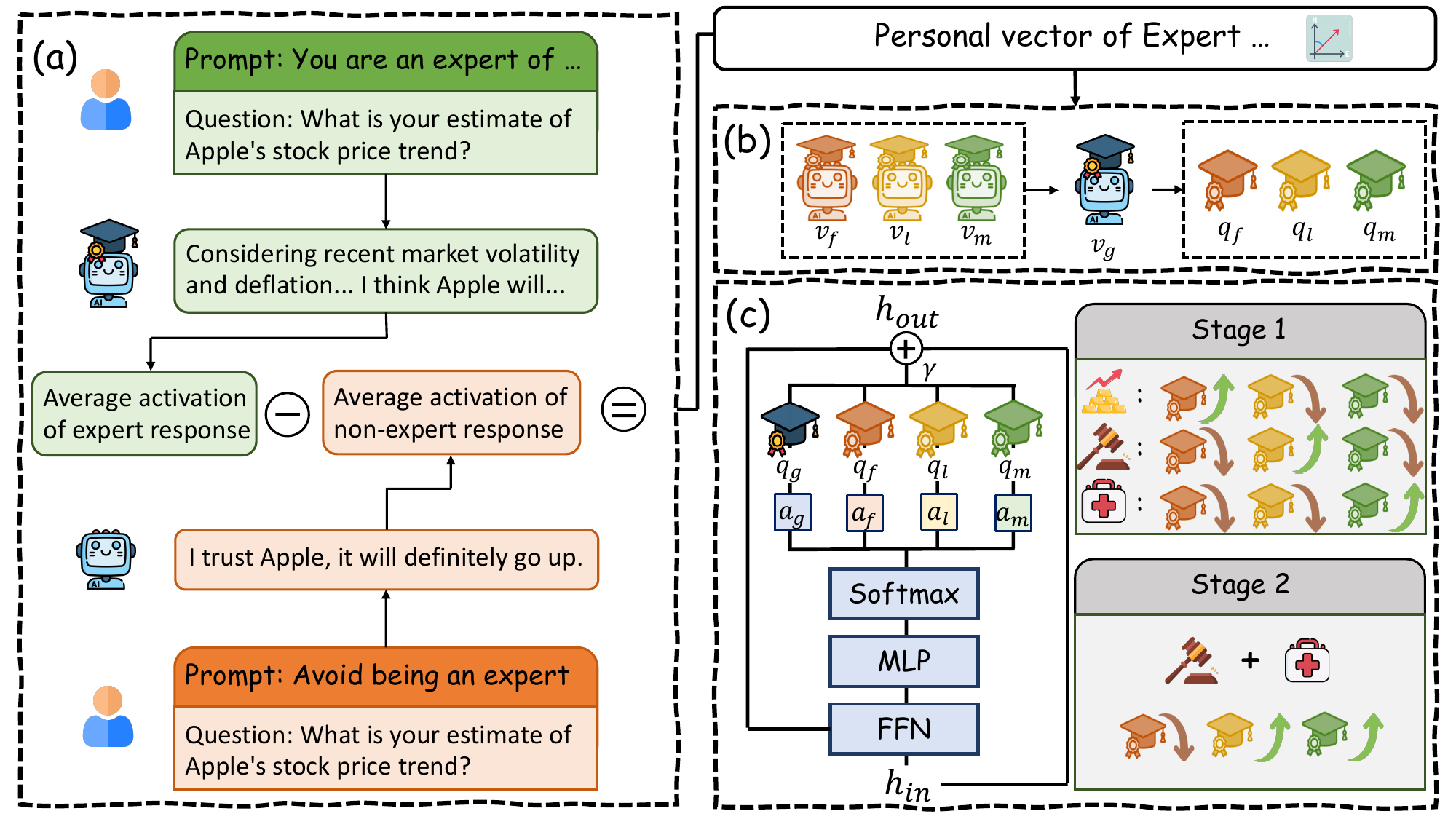} 
\caption{Our FOCUS algorithm workflow. (a) Extract general expert and single-domain expert persona vectors. (b) Decouple single-domain expert vectors from the general expert vector. (c) Conduct two-stage fine-tuning to train the expert gating module, where the first stage aligns single-domain expert personas and the second stage enhances cross-domain generalization.}
\label{fig:focus}
\end{figure}

\subsection{Expert Persona Vector Extraction}
We first design an expert persona vector extraction pipeline, inspired by~\cite{chen2025persona}. This pipeline takes as input the names of domain experts, corresponding brief descriptions of expert traits, and question-answer samples extracted from domain training sets. Based on these inputs, we design a universal prompt template to guide a state-of-the-art LLM (Claude 3.7 Sonnet) to generate two types of key components: contrastive system prompts and evaluation prompts.

Specifically, for each domain, we generate 5 pairs of contrastive system prompts. Each pair includes a positive prompt designed to elicit target expert persona behaviors and a negative prompt designed to suppress target expert persona behaviors. We select 20 questions from each domain's training set, which, when combined with the contrastive system prompts, can effectively guide the target model to produce behaviors related to specific expert personas.

Using the above components, we construct contrastive pairs of model responses. For each question in the extraction set, we use positive and negative prompts to drive the target model to generate 10 responses, respectively. 
For each filtered response $r$, we extract its residual stream activation at layer $l$ of the model and average across all token dimensions to obtain the average activation vector $\bar{\mathbf{a}}_l(r)$. Finally, as shown in Fig.~\ref{fig:focus}(a), the persona vector $\mathbf{v}^{(l)}$ at layer $l$ is defined as the difference between the average activation vectors of positive and negative response sets:

\begin{equation}
\mathbf{v}^{(l)} = \frac{1}{|S_{pos}|} \sum_{r \in S_{pos}} \bar{\mathbf{a}}_l(r) - \frac{1}{|S_{neg}|} \sum_{r \in S_{neg}} \bar{\mathbf{a}}_l(r),
\end{equation}

where $S_{pos}$ and $S_{neg}$ represent the sets of responses generated by positive and negative prompts, respectively. For simplicity, in subsequent discussions, we omit the layer index $l$ and represent the candidate expert persona vectors extracted from specific layers for finance, legal, and medical domains as $\mathbf{v}_f, \mathbf{v}_l, \mathbf{v}_m$, respectively. Meanwhile, we use more general prompts and mixed-domain questions to construct a general expert persona vector $\mathbf{v}_g$ through the same process.

\subsection{Expert Persona Vector Decoupling}
\label{sec:decoupling}
Although candidate expert persona vectors can already be used for training-free activation or for fine-tuning~\cite{chen2025persona}, domain expert persona vectors directly extracted from models are formed through pre-training and post-training data, which may contain cross-domain coupling data that interferes with the expression of pure expert personas for specific domains. To address this issue, we employ the Gram-Schmidt orthogonalization method to decouple the vectors, aiming to remove general expert components from each domain expert vector.

Specifically, we first use the general expert vector $\mathbf{v}_g$ as a reference and normalize it to obtain the orthogonal basis vector $\mathbf{q}_g$:
$\mathbf{q}_g = \frac{\mathbf{v}_g}{|\mathbf{v}_g|}$.

Subsequently, for each domain expert vector $\mathbf{v}_k$ (where $k \in {f, l, m}$), we subtract its projection component in the $\mathbf{q}_g$ direction to obtain a vector $\mathbf{v}'_k$ orthogonal to the general expert component:

\begin{equation}
\mathbf{v}'_k = \mathbf{v}_k - (\mathbf{q}_g^T \mathbf{v}_k)\mathbf{q}_g.
\end{equation}

Finally, we normalize $\mathbf{v}'_k$ to obtain the decoupled domain expert persona vector $\mathbf{q}_k$:
$\mathbf{q}_k = \frac{\mathbf{v}'_k}{|\mathbf{v}'_k|}$.

Through this process, we obtain a set of decoupled orthogonal persona vectors ${\mathbf{q}_g, \mathbf{q}_f, \mathbf{q}_l, \mathbf{q}_m}$, where each domain expert vector $\mathbf{q}_f, \mathbf{q}_l, \mathbf{q}_m$ is orthogonal to the general expert vector $\mathbf{q}_g$.

\subsection{Two-stage Training Expert Gating}
After obtaining decoupled expert vectors, the core challenge lies in how to guide the model to dynamically activate one or more expert personas based on context. To this end, we introduce a lightweight expert gating module, as shown in Fig.~\ref{fig:focus}(c). This module is deployed after the feed-forward network (FFN) sublayer of the Transformer block. Let the FFN output be $\mathbf{h}_{\text{ffn}} \in \mathbb{R}^d$, and the expert persona vector set be ${\mathbf{q}_g, \mathbf{q}_f, \mathbf{q}_l, \mathbf{q}_m} \subset \mathbb{R}^d$. The expert gating module consists of a two-layer MLP $g(\cdot)$ and a Softmax function.

First, the module computes unnormalized logits $\mathbf{z}$ based on the FFN output: $\mathbf{z} = g(\mathbf{h}_{\text{ffn}}) \in \mathbb{R}^4$. Subsequently, normalized weight vector $\mathbf{\alpha}$ is obtained through Softmax:

\begin{equation}
\mathbf{\alpha} = \text{softmax}(\mathbf{z}) = [\alpha_g, \alpha_f, \alpha_l, \alpha_m],
\end{equation}

where $\sum_{k \in {g,f,l,m}} \alpha_k = 1$ and $\alpha_k \ge 0$. Using these weights, we construct a weighted persona offset vector $\mathbf{r}$:

\begin{equation}
\mathbf{r} = \alpha_g \mathbf{q}_g + \alpha_f \mathbf{q}_f + \alpha_l \mathbf{q}_l + \alpha_m \mathbf{q}_m.
\end{equation}

Finally, this persona offset vector is added to the residual connection through a scaling factor $\gamma$. Let the input to the FFN sublayer be $\mathbf{h}_{\text{in}}$, then the final output $\mathbf{h}_{\text{out}}$ of this Transformer block is computed as follows:

\begin{equation}
\mathbf{h}_{\text{out}} = \mathbf{h}_{\text{in}} + \mathbf{h}_{\text{ffn}} + \gamma \cdot \mathbf{r},
\end{equation}

where $\gamma$ is a hyperparameter for controlling the injection strength of persona vectors. Normalization operations are omitted in the formula.

Ideally, for single-domain tasks, the model should only activate the corresponding expert vector; for cross-domain tasks, it should activate multiple relevant expert vectors. To achieve this goal, we design a two-stage fine-tuning strategy.

\textbf{Stage 1}: Single-Domain Alignment and Sparsity Regularization.
We use single-domain datasets $D_f, D_l, D_m$ that exclude cross-domain samples to fine-tune the model in stages. Each sample $x$ comes with its domain label $k \in {f, l, m}$. The training objective of this stage is to promote the gating weights $\mathbf{\alpha}$ to exhibit a sparse unimodal distribution of $\alpha_k \rightarrow 1$ when processing tasks in domain $k$, while ensuring model generation quality. To this end, we introduce a gated selection regularizer term in the loss function:

\begin{equation}
\mathcal{L}_{\text{total}} = \mathcal{L}_{\text{LM}} + \lambda \cdot \text{CE}(\text{Onehot}(k), \mathbf{\alpha}),
\end{equation}
where the first term is the standard language model cross-entropy loss. The second term is the gated selection regularizer term, defined as the cross-entropy loss between the gating weights $\mathbf{\alpha}$ and the one-hot encoding of the corresponding domain label. $\lambda$ ($\lambda=0.2$) is a hyperparameter for balancing the two loss terms.

\textbf{Stage 2}: Cross-Domain Generalization and Adaptive Learning. 
We use a new dataset with the same data sources as \textbf{Stage 1} but non-overlapping samples, which contains a mixture of single-domain and cross-domain samples. During training, we remove the gated selection regularizer term from the loss function, the total loss is then defined as:

\begin{equation}
\mathcal{L}_{\text{total}} = \mathcal{L}_{\text{LM}} .
\end{equation}
The objective of this stage is to guide the model to adaptively learn the allocation of gating weights $\mathbf{\alpha}$ based on input context without imposing explicit selection constraints, thereby mastering reasonable weight combinations in cross-domain scenarios and improving model performance and robustness on complex tasks.

\section{Experiments}

\textbf{Datasets}. For the finance domain, CoT-stock-2k~\cite{chu-etal-2025-domaino1s} is used for training, and the stock prediction dataset provided by Koa et al.~\cite{koa2024learning} is used for evaluation. For the legal domain, CoT-legal-2k~\cite{chu-etal-2025-domaino1s} is used for training, and LegalBench~\cite{guha2023legalbench} is used for evaluation. For the medical domain, 2k questions selected from MedQA~\cite{jin2021disease} are used for training, and the professional medicine task from MMLU~\cite{hendrycks2020measuring} is used for evaluation. For cross-domain evaluation, the legal-medical cross-domain dataset MedEthicsQA~\cite{wei2025medethicsqa} and the finance-legal cross-domain dataset CUAD~\cite{hendrycks2021cuad} are used. For our FOCUS, each of the two-stage fine-tuning processes uses half of the corresponding training set for training.

\textbf{Baseline Methods}. (I) Baselines for LLM persona control. Prompt~\cite{jiang-etal-2024-personallm}: activates corresponding expert personas through prompt learning. Inference-Time Steering~\cite{chen2025persona}: directly adds extracted expert persona vectors to FFN layers during inference. SFT: supervises fine-tuning of the model to generate responses in the manner of domain experts, training on manually-checked responses to questions generated by GPT-4o. CAFT~\cite{casademunt2025steering} and Preventative Steering~\cite{chen2025persona}: through ablating certain personas (in this paper, the opposite of expert personas, i.e., non-expert personas) using linear projections during fine-tuning, thereby enabling the model to retain the desired expert personas. (II) General and domain LLMs. Qwen-2.5-7B-Instruct~\cite{qwen2.5} and Llama-3-8B-Instruct~\cite{llama3modelcard} serve as baselines for general LLMs. For finance domain LLMs, Finance-Chat~\cite{cheng2023adapting} and Domaino1s-finance~\cite{chu-etal-2025-domaino1s} are selected. For legal domain LLMs, Law-Chat~\cite{cheng2023adapting} and Domaino1s-legal~\cite{chu-etal-2025-domaino1s} are selected. For medical domain LLMs, UltraMedical-LLama3-7B~\cite{zhang2024ultramedical} and BioMistral-7B~\cite{labrak2024biomistral} are selected.

\textbf{Implementation Details}. 
The learning rate, training epochs, global batch size, and maximum sequence length are set to 2e-5, 30, 16, and 4096, respectively. The hyperparameter $\gamma$ is set to 2. All experiments run on 4 NVIDIA A800-SXM4 80G GPUs. All operations targeting expert persona vectors are implemented at layer 20 of the LLM, consistent with~\cite{chen2025persona}.

\subsection{Main Results}

As shown in Table~\ref{tab:focus_comparison}, Qwen-FOCUS and Llama-FOCUS achieve strong performance across financial, legal, medical, and cross-domain benchmarks. Compared with general-purpose base models, FOCUS consistently improves accuracy on most evaluation tasks, showing that explicitly controlling expert personas is beneficial for domain-specific reasoning. Compared with standard SFT, FOCUS also achieves better or comparable results across most tasks, suggesting that the proposed persona-vector control mechanism provides additional benefits beyond supervised fine-tuning alone.

\begin{table*}[htbp]
\renewcommand{\arraystretch}{1.2}
\resizebox{\textwidth}{!}{%
\begin{tabular}{l|c|cccccccc|c|cc}
\toprule
\multicolumn{1}{c}{\multirow{2}{*}{\textbf{Method}}} & \multicolumn{1}{c}{\textbf{Finance}} & \multicolumn{8}{c}{\textbf{Legal}}  & \multicolumn{1}{c}{\textbf{Medical}} & \multicolumn{2}{c}{\textbf{Cross-Domain}} \\
\cmidrule{2-13}
\multicolumn{1}{c}{} & Stock Predict & CC & IP & MAUD & PJ & PP & scalr & TTD & TTP & Professional Medicine & MedEthicsQA & CUAD \\
\midrule
\multicolumn{13}{c}{\textit{Baselines for activating LLM personas}} \\
\midrule
Prompt & 48.62 & 27.55 & 41.35 & 33.81 & 56.00 & 28.83 & 68.83 & 76.64 & 48.16 & 75.36 & 72.84 & 57.96 \\
Inference Steering & 44.78 & 34.09 & 44.36 & 32.29 & 48.00 & 21.17 & 68.83 & 76.64 & 49.70 & 68.01 & 71.11 & 52.50 \\
SFT & 51.33 & 78.03 & 40.06 & 60.29 & 70.00 & 56.39 & 65.50 & 98.13 & 87.88 & 75.60 & 79.79 & 81.05 \\
CAFT & 51.23 & 77.02 & 47.37 & 55.21 & 65.00 & 53.20 & 62.11 & 97.22 & 81.04 & 73.16 & 74.55 & 78.10 \\
Preventative Steering & 52.33 & 76.11 & 47.37 & 57.86 & 64.00 & 54.40 & 61.12 & 98.13 & 83.03 & 73.53 & 75.28 & 80.80 \\
\midrule
\multicolumn{13}{c}{\textit{Baselines for general and domain LLMs}} \\
\midrule
Qwen-2.5-7B-Instruct & 50.74 & 27.02 & 42.86 & 34.45 & 66.00 & 28.45 & 68.83 & 79.44 & 47.88 & 77.94 & 73.24 & 58.22 \\
Llama-3-8B-Instruct & 48.12 & 24.85 & 40.21 & 31.78 & 62.50 & 26.12 & 65.47 & 76.89 & 45.23 & 74.25 & 71.34 & 55.67 \\
Finance-Chat & 49.85 & - & - & - & - & - & - & - & - & - & - & 58.63 \\
Domain-o1s-stock & 51.92 & - & - & - & - & - & - & - & - & - & - & - \\
Law-Chat & - & 26.78 & 41.53 & 33.21 & 64.00 & 27.89 & 67.25 & 78.12 & 46.55 & - & - & 57.43 \\
Domain-o1s-legal & - & 28.34 & 43.17 & 35.67 & 67.50 & 29.73 & 69.58 & 80.27 & 48.91 & - & - & 80.88 \\
UltraMedical-7B & - & - & - & - & - & - & - & - & - & \textbf{84.19} & 78.23 & - \\
BioMistral-7B & - & - & - & - & - & - & - & - & - & 55.14 & 66.87 & - \\
\midrule
\textbf{Llama-FOCUS} & 51.76 & 86.12 & 48.12 & 60.62 & 70.00 & \textbf{58.25} & 69.75 & 97.19 & 88.48 & 76.70 & 80.49 & 82.82 \\
\textbf{Qwen-FOCUS} & \textbf{52.47} & \textbf{86.38} & \textbf{49.62} & \textbf{64.20} & \textbf{72.00} & 58.10 & \textbf{70.22} & \textbf{98.13} & \textbf{89.09} & 79.04 & \textbf{81.61} & \textbf{83.15} \\
\bottomrule
\end{tabular}%
}
\vspace{1pt}
\caption{Performance comparison of FOCUS with other methods on finance, legal, medical, and cross-domain tasks. Baselines for activating LLM personas are all based on Qwen-2.5-7B-Instruct as the foundation model.}
\label{tab:focus_comparison}
\end{table*}

In the finance and legal domains, FOCUS shows particularly clear advantages. On the stock prediction task, Qwen-FOCUS achieves the best accuracy among all compared methods. In the legal domain, Qwen-FOCUS obtains strong improvements across most LegalBench subsets, including contract-related and rule-based reasoning tasks. These results indicate that decoupled expert personas can better capture domain-specific reasoning patterns and reduce irrelevant behavioral interference. Notably, FOCUS also outperforms several domain-specific LLMs, such as Finance-Chat and Law-Chat, which rely on domain-specific training. This suggests that FOCUS can serve as an efficient alternative for domain adaptation by directly controlling internal expert persona representations rather than relying only on large-scale domain data.

FOCUS also demonstrates strong performance on cross-domain benchmarks. On MedEthicsQA and CUAD, Qwen-FOCUS achieves the best results among all compared methods. These tasks require the model to integrate multiple types of expertise, such as medical and legal reasoning, rather than simply activating a single domain-specific behavior. The improvement on these benchmarks suggests that the expert gating module can adaptively compose multiple persona vectors according to the input context. This is consistent with our motivation that expert personas should be decoupled and selectively activated, especially when different domains impose different reasoning requirements.

In contrast, direct persona activation methods, including prompting and inference-time steering, show less stable performance and may underperform the base model or SFT baseline. This suggests that simply activating an entangled expert persona is insufficient. Overall, the results support the effectiveness of FOCUS in improving both single-domain specialization and cross-domain generalization.

\subsection{Ablation Experiments}
\subsubsection{Effects of the Expert Gating Module}

\begin{figure}[htbp]
\centering
\includegraphics[width=0.95\textwidth]{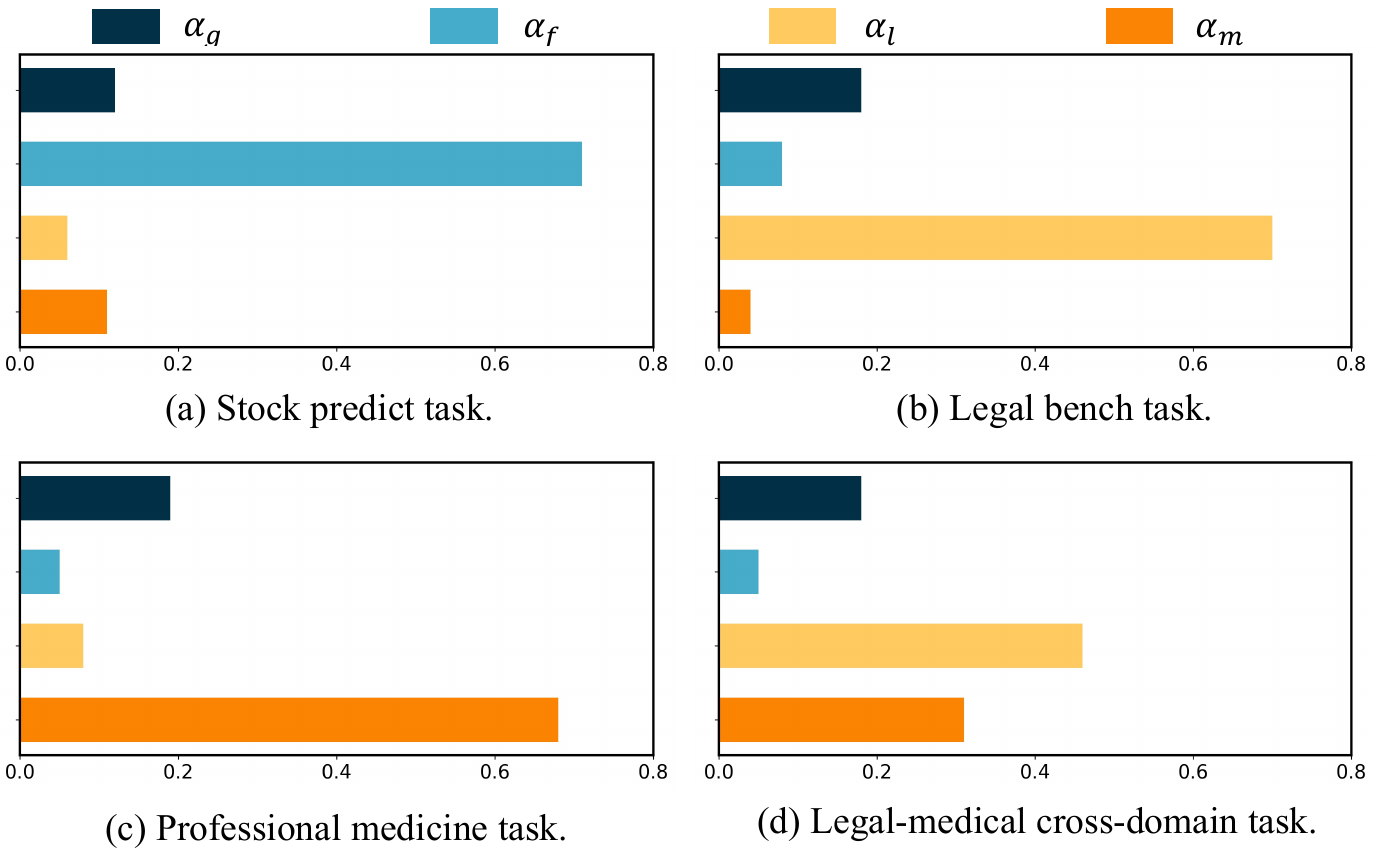} 
\caption{Distribution of the gating weight vector $\mathbf{\alpha}$.}
\label{fig:exp2}
\end{figure}

As shown in Fig.~\ref{fig:exp2}(a), (b), and (c), when the model processes purely financial, legal, or medical tasks, the expert gating module primarily activates the corresponding domain expert vectors, exhibiting sparsity and selectivity. This indicates that the model learns to precisely invoke the corresponding expert personas for domain-specific tasks. When processing the medical-legal cross-domain MedEthicsQA task, as shown in Fig.~\ref{fig:exp2}(d), the model instead assigns high weights to both the medical expert vector $\mathbf{q}_m$ and the legal expert vector $\mathbf{q}_l$. This demonstrates that the model can identify the composite attributes of the task and dynamically fuse multiple expert personas to address complex cross-domain challenges.

\subsubsection{Necessity of the Two-Stage Training Strategy} 
We evaluate the necessity of the two-stage training strategy on the Professional Medicine dataset. As shown in Table~\ref{tab:ablation_two_stage}, removing Stage 1 leads to performance degradation, indicating that single-domain alignment is important for learning focused expert activation. Without this stage, the model struggles to associate each domain input with the corresponding expert persona vector. Removing Stage 2 also reduces performance, suggesting that cross-domain adaptive learning helps the model combine multiple expert personas under complex contexts.

\begin{table}[htbp]
\centering
\small
\renewcommand{\arraystretch}{1.2}
\begin{tabular}{lccc}
\toprule
Method & w/o Stage 1 & w/o Stage 2 & \textbf{Qwen-FOCUS} \\
\midrule
Acc(\%) & 76.17 & 78.69 & \textbf{79.04} \\
\bottomrule
\end{tabular}
\vspace{3pt}
\caption{Ablation study of two-stage training, w/o indicates that the corresponding training stage is removed.}
\label{tab:ablation_two_stage}
\end{table}

These results show that the two stages play complementary roles: Stage 1 provides a reliable foundation by learning focused, domain-specific expert activations, while Stage 2 enables more flexible persona combinations. Therefore, both stages are necessary for stable performance.


\subsubsection{Impact of $\gamma$.}

We also explore the impact of $\gamma$ on model performance, taking the Professional Medicine task as an example. As shown in Table~\ref{tab:gamma_impact}, when $\gamma=0$, model performance declines. When $\gamma$ takes very large values (such as 5 and 10), model performance also shows a decline. We refer to this as “persona over-reinforcement”. 
Excessively large $\gamma$ may lead the model to overfit expert characteristics and lose generalization ability, or even force the display of certain personas in inappropriate scenarios, resulting in incorrect answers. Finally, we adopt the setting of $\gamma=2$, which demonstrates good performance across most tasks.

\begin{table}[htbp]
\centering
\small
\renewcommand{\arraystretch}{1.2}
\begin{tabular}{cc|cc}
\toprule
\textbf{$\gamma$} &\textbf{ Accuracy(\%)} & \textbf{$\gamma$} & \textbf{Accuracy(\%)} \\
\midrule
$\gamma=0$ & 77.82 & $\gamma=3$ & 78.92 \\
$\gamma=1$ & 78.82 & $\gamma=5$ & 72.03 \\
$\gamma=2$ & \textbf{79.04} & $\gamma=10$ & 70.95 \\
\bottomrule
\end{tabular}
\vspace{3pt}
\caption{Impact of $\gamma$ on accuracy (on Qwen-FOCUS).}
\label{tab:gamma_impact}
\end{table}

\subsubsection{Effect of Persona Vector Decoupling}

To evaluate the necessity of the orthogonal decoupling step introduced in Section~\ref{sec:decoupling}, we conduct an ablation study by removing the decoupling operation. Specifically, instead of projecting each domain-specific persona vector onto the orthogonal subspace of the general expert vector, we directly use the original extracted vectors $v_f$, $v_l$, and $v_m$ together with the general vector $v_g$ as inputs to the gating module. All other settings remain unchanged, including the two-stage training strategy, the gating architecture, and the hyperparameter setting.

As shown in Table~\ref{tab:decoupling_ablation}, removing the decoupling step leads to consistent performance degradation on both single-domain and cross-domain tasks. In particular, the model shows lower accuracy on financial, legal, and medical benchmarks, indicating that the raw persona vectors are less effective for representing clean domain-specific expertise. We attribute this to the residual general-expert and cross-domain components embedded in the original vectors, which introduce interference during persona activation.

A similar trend is also observed on the cross-domain benchmarks MedEthicsQA and CUAD. This suggests that, without decoupling, the model is less capable of composing multiple expert personas in a precise and complementary manner. Instead, the entangled persona representations make the routing process less selective and weaken the model's ability to adapt to complex mixed-domain scenarios.

Overall, these results demonstrate that orthogonal decoupling is an important component of FOCUS. By removing shared and interfering components from domain-specific persona vectors, it improves the purity of expert representations and facilitates more accurate and controllable expert activation.

\begin{table*}[h]
\vspace{3pt}
\renewcommand{\arraystretch}{1.2}
\resizebox{\textwidth}{!}{%
\begin{tabular}{l|c|cccccccc|c|cc}
\toprule
\multicolumn{1}{c}{\multirow{2}{*}{\textbf{Method}}} & \multicolumn{1}{c}{\textbf{Finance}} & \multicolumn{8}{c}{\textbf{Legal}}  & \multicolumn{1}{c}{\textbf{Medical}} & \multicolumn{2}{c}{\textbf{Cross-Domain}} \\
\cmidrule{2-13}
\multicolumn{1}{c}{} & Stock Predict & CC & IP & MAUD & PJ & PP & scalr & TTD & TTP & Professional Medicine & MedEthicsQA & CUAD \\
\midrule
w/o decoupling & 51.21 & 84.72 & 48.01 & 62.48 & 70.50 & 56.91 & 69.37 & 97.22 & 87.96 & 77.96 & 79.88 & 81.34 \\
\textbf{Qwen-FOCUS} & \textbf{52.47} & \textbf{86.38} & \textbf{49.62} & \textbf{64.20} & \textbf{72.00} & \textbf{58.10} & \textbf{70.22} & \textbf{98.13} & \textbf{89.09} & \textbf{79.04} & \textbf{81.61} & \textbf{83.15} \\
\bottomrule
\end{tabular}%
}
\vspace{3pt}
\caption{Ablation study on persona vector decoupling. ``w/o decoupling'' directly uses the original persona vectors without orthogonal decomposition. The results show that decoupling reduces cross-domain interference and improves domain-specific persona purity.}
\label{tab:decoupling_ablation}
\end{table*}

\section{Conclusion}

In this paper, we propose FOCUS, a persona-vector editing method for decoupling and controlling domain-specific expert personas in LLMs. FOCUS applies orthogonal decomposition to reduce cross-domain persona coupling and uses a lightweight expert gating module to adaptively activate suitable persona vectors according to task contexts.

Experiments on financial, legal, medical, and cross-domain benchmarks show that FOCUS consistently improves accuracy over existing persona activation methods, standard fine-tuning, and domain-specific models. Gating analysis and ablation studies further verify the roles of orthogonal decoupling, two-stage training, and controlled persona injection, highlighting precise persona control as a promising direction for efficient LLM domain adaptation.

%
%
\bibliographystyle{splncs04}
\bibliography{refs}

\end{document}